\documentclass[12pt]{article}

\usepackage[margin=2.0cm]{geometry}
\usepackage{amsmath}

\usepackage{multirow}
\usepackage{enumitem}
\usepackage{amssymb}
\usepackage{amsthm}
\usepackage{amsmath}
\usepackage{graphicx}
\usepackage{array} 
\usepackage{subcaption}
\usepackage{lipsum} 
\usepackage{setspace}
\usepackage[sort&compress,numbers]{natbib}
\usepackage[colorlinks=true, linkcolor=blue, citecolor=blue, urlcolor=blue]{hyperref} 

\usepackage{multirow}
\usepackage{booktabs}
\usepackage{tabularx}
\graphicspath{ {figures/} }
\usepackage{float}

\usepackage{tikz}
\usetikzlibrary{shapes.geometric, arrows.meta, positioning}
\usepackage{algorithm}
\usepackage{algorithmic}
\usepackage{pdflscape}
\usepackage{graphicx}

\usepackage{tikz}
\usetikzlibrary{shapes.geometric, arrows, positioning}

\tikzstyle{process} = [rectangle, minimum width=3cm, minimum height=1cm, text centered, draw=black]

\title{Machine Learning-Enhanced Tabu Search for Tactical Wireless Network Design}

\author{
	Wissem Ahmed Zaid,
	Alain Hertz,
    Defeng Liu
	\\[3mm]
	\footnotesize Polytechnique Montr\'eal - GERAD, Montr\'eal, Canada\\[3mm]
}

\date{\today}

\begin{document}

\maketitle

\begin{abstract} 

Designing high-performance tactical wireless networks under realistic operational constraints gives rise to challenging combinatorial optimization problems, where the evaluation of candidate solutions relies on detailed physical and traffic-aware models. Although classical metaheuristics such as Tabu Search offer effective mechanisms for exploring large search spaces, their computational cost remains high because numerous candidate moves must be evaluated at every iteration. In this paper, we propose a data-driven framework that improves the efficiency of Tabu Search by learning to guide its move selection process. Rather than altering the neighborhood structure, our approach exploits the information contained in the search trajectories generated during the optimization process. At each iteration, we record both improving and non-improving edge-based transformations together with a set of descriptive features capturing the structural, geometric, and performance characteristics of the network. This information is used to train a Graph Neural Network (GNN) that predicts the impact of candidate moves on the objective function. The trained model is then integrated into the Tabu Search algorithm to rank candidate transformations according to their predicted quality, thereby reducing the number of costly objective evaluations while maintaining an effective exploration of the search space. Experimental results on synthetic benchmark instances demonstrate that the proposed learning-assisted Tabu Search notably reduces computation time while consistently producing higher-quality solutions than the standard algorithm. These findings highlight the potential of combining machine learning with metaheuristics by leveraging the implicit knowledge embedded in search trajectories, paving the way for more efficient solution methods for large-scale network design problems.\end{abstract}

\vspace{0.3cm}\noindent \emph{Keywords:} tactical wireless network design; machine learning; metaheuristic; graph neural network.

\section{Introduction}

Wireless communication systems play a critical role in environments where conventional telecommunication infrastructures are unavailable, damaged, or inadequate. This is particularly true in emergency response, military operations, and remote-area deployments, where temporary tactical wireless networks must be rapidly deployed to ensure reliable communication among geographically distributed locations.

The design of such networks naturally gives rise to a challenging combinatorial optimization problem. Several interdependent decisions must be made, including the selection of a central coordinating node, the construction of a feasible network topology, and the configuration of communication links and radio resources. These decisions have a significant impact on network performance, as they directly influence signal quality, interference levels, and effective throughput.

Metaheuristic approaches, such as Tabu Search, provide a flexible and effective framework for tackling this class of optimization problems. They are particularly well suited to instances where exact methods become impractical due to the vast search space and the high computational cost of evaluating candidate solutions. However, the efficiency of a Tabu Search algorithm depends critically on how its neighborhood is explored. In the tactical wireless network design problem, each iteration may involve a large number of candidate topological modifications, and exhaustively evaluating all of them can be computationally prohibitive.

This observation motivates the integration of machine learning into the search process. Rather than replacing the optimization algorithm, the learning component is designed to guide neighborhood exploration by identifying candidate moves more likely to produce high-quality solutions. In this way, machine learning serves as a decision-support mechanism within the metaheuristic, enabling the algorithm to focus on the most promising regions of the search space while preserving the feasibility constraints and evaluation procedures of the original optimization framework.

In this paper, we propose a machine learning-guided Tabu Search algorithm for tactical wireless network design. The proposed approach learns from previously generated search trajectories to prioritize candidate neighborhood moves before their computationally expensive evaluation. The objective is to improve the efficiency of the search while maintaining, and potentially enhancing, the quality of the solutions obtained.

This work makes several contributions to the tactical wireless network design problem. We propose a machine learning-guided optimization framework that integrates a learning component within the search process to improve the efficiency of the optimization procedure. We also develop a graph-based learning model that exploits both structural characteristics of the network and solution-dependent features to identify promising neighborhood moves. Finally, we perform an extensive computational study to evaluate the effectiveness of the proposed approach and analyze the impact of the learning component on solution quality, search efficiency, and performance.

The remainder of the paper is organized as follows. Section~\ref{sec:problem_description} presents the tactical wireless network design problem. Section~\ref{sec:literature_review} reviews the related literature on wireless network design, metaheuristics, and machine learning for combinatorial optimization. Section~\ref{sec:topology_tabu_search} describes the baseline Topology Tabu Search algorithm, including its edge-exchange neighborhood structure. Section~\ref{sec:tabu_ml} introduces the proposed machine learning-guided framework. It describes the learning-guided move selection strategy, the generation of the training dataset, and the design of the Graph Neural Network (GNN)-based edge classifier, including the feature representation, network architecture, and training objective. Section~\ref{sec:ml_guided_tabu} then presents the complete ML-guided Tabu Search algorithm. Computational experiments and performance analyses are reported in Section~\ref{sec:computational_experiments}. Finally, Section~\ref{sec:conclusion} concludes the paper and outlines directions for future research.

\section{Problem description}\label{sec:problem_description}

A detailed description of the complete tactical wireless network design problem is provided in \cite{WH1}. In this section, we briefly review the main elements required to understand the proposed learning-guided Tabu Search framework.

An instance of the problem is defined by a set of $n$ nodes, each associated with fixed geographical coordinates. The network must be constructed as a tree topology connecting all nodes. Once a tree is selected, one node is designated as the master hub, and the tree is oriented outward from this root node. The master hub acts as the central coordinating node, while each remaining node has exactly one predecessor and may have multiple successors.

Each node is equipped with a radio interface connected to two multi-beam antennas. The radio operates over two channels, each of which can be assigned one of two available transmission frequencies. Communication links can operate either in point-to-point (PTP) mode, where a node communicates with a single successor, or in point-to-multipoint (PMP) mode, where a node simultaneously serves multiple successors. These design choices, combined with the antenna configuration and channel/frequency assignment, determine the signal quality, inter-link interference, and ultimately the achievable network throughput.

\begin{figure}[!htb]
    \centering
    \includegraphics[scale=0.5]{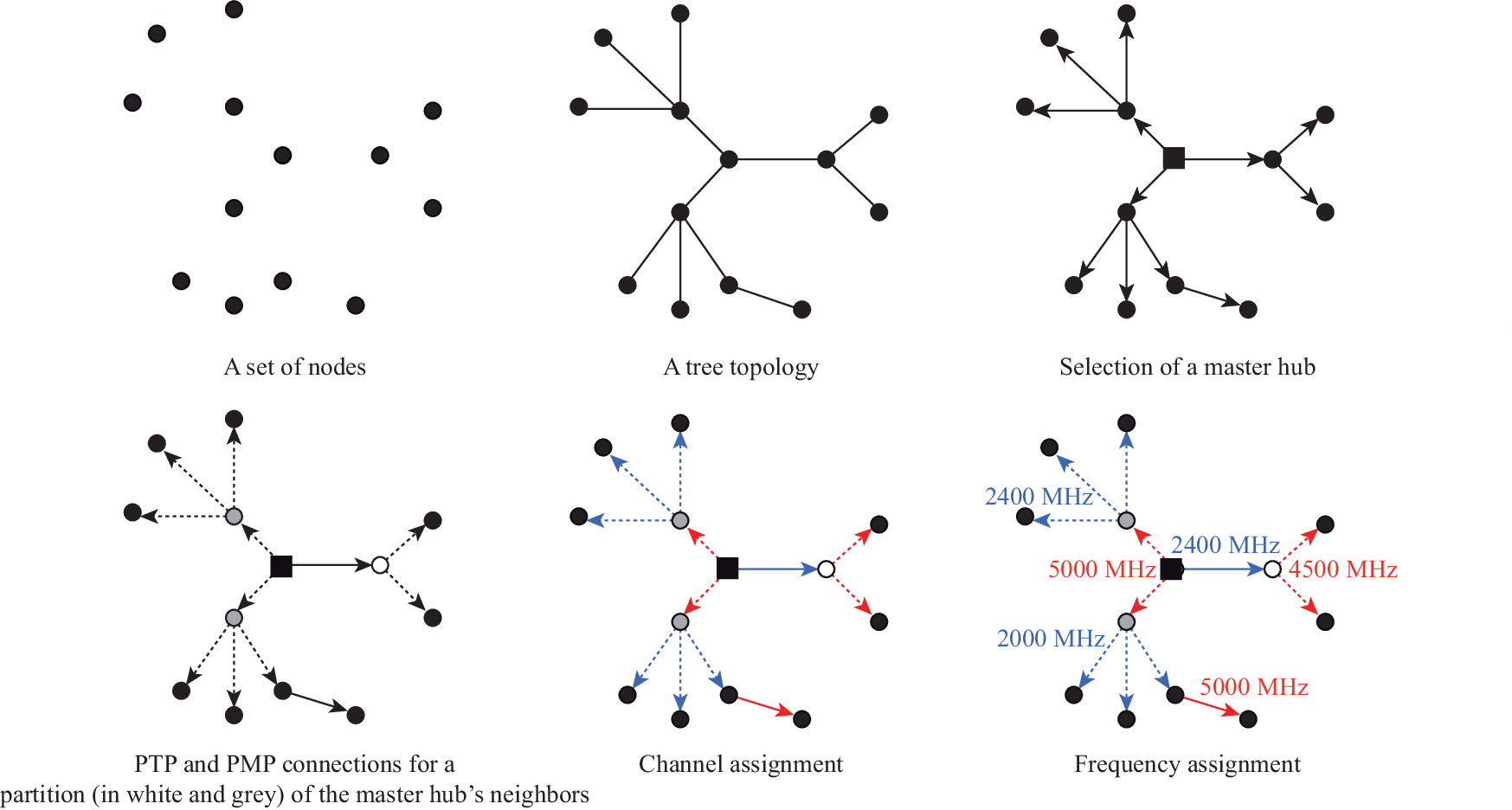}
    \caption{Illustration of the tactical wireless network design process, reproduced from \cite{WH1}.}

    \label{fig:process}
\end{figure}

Figure \ref{fig:process}, reproduced from \cite{WH1}, illustrates the main steps of the network design process. Given an initial set of nodes, a tree topology is first constructed to establish connectivity among all nodes. One node is then selected as the master hub, represented by a black square in the figure. The nodes directly connected to the master hub are subsequently partitioned into two groups: a group of grey nodes (two nodes in the illustrated example) and a group of white nodes (one node in the illustrated example).
The figure also highlights the different link configurations: solid lines represent point-to-point (PTP) connections, while dashed lines represent point-to-multipoint (PMP) connections. Communication channels are then assigned, with the two channels depicted in red and blue, respectively. Finally, transmission frequencies are allocated to each channel; in the example, the red channel uses frequencies of 4500 MHz and 5000 MHz, whereas the blue channel uses frequencies of 2000 MHz and 2400 MHz.
The figure does not display the activated antenna beams or their exact orientations, which are determined separately through a geometric procedure.

At this stage, it is important to distinguish between a {\it topology}, a {\it configuration}, and the resulting {\it solution}. A topology $T$ specifies only the connectivity of the network, that is, which pairs of nodes are linked by an edge. Before its performance can be evaluated, a number of additional configuration decisions must be made.
The following examples illustrate the nature of these decisions. One of the $n$ nodes must first be selected as the master hub. If the chosen hub has $x$ neighbors, these neighbors can be partitioned into two groups in $2^{x-1}$ different ways. In addition, one of the two available frequencies must be assigned to each edge, yielding $2^{n-1}$ possible frequency assignments. These are only a few examples of the decisions involved in the configuration process, and they are strongly interdependent. Consequently, a single topology may admit an extremely large number of feasible configurations.

The procedure proposed in \cite{WH1} constructs one such configuration by combining several heuristics that restrict the search to a limited set of promising candidates. Although the resulting configuration is generally of good quality, optimality is not guaranteed for the given topology. In the remainder of the paper, we denote by $s(T)$ the solution produced by this heuristic configuration procedure when applied to topology $T$.

Ideally, the quality of a topology would be defined by the objective value of its best feasible configuration. Since exhaustively evaluating all configurations induced by a topology is computationally intractable, we instead assess a topology through the heuristic solution $s(T)$ returned by the above procedure. We now define the objective function $f(s)$ used to evaluate any solution $s$.

Let $s$ be a solution with edge set $E$.
For each edge $uv\in E$, the direct throughput $TP_{uv}$ is computed using a physical-layer model that accounts for both signal characteristics and interference between edges operating on the same frequency. Further details on this computation are provided in \cite{WH1}. Three traffic scenarios are considered. Let $n^X_{uv}$ denote the number of flows carried by edge $uv$ under scenario $X$, and let $d_v$ denote the number of descendants of node $v$ in the rooted directed tree. In scenario $A$, a single flow is associated with each edge, leading to $n^A_{uv}=1$. Scenario $B$ corresponds to simultaneous communication between the master hub and all other nodes; in this case, $n^B_{uv}=d_v$ if the edge is directed from $u$ to $v$, and $n^B_{uv}=d_u$ otherwise. Finally, scenario $C$ considers bidirectional communication between every pair of nodes. The resulting edge load is $n^C_{uv}=2d_v(|V|-d_v)$ when the edge is directed from $u$ to $v$, and $n^C_{uv}=2d_u(|V|-d_u)$ otherwise.
The objective is to maximize a weighted combination of the minimum and average effective throughput over the three scenarios. Specifically, with $p$ controlling the trade-off between these two criteria and $\omega_X$ denoting the weight associated with scenario $X\in\{A,B,C\}$, the objective function $f(s)$ is defined as:
\[
f(s)=\sum_{X\in\{A,B,C\}} \omega_X
\left(
\min_{uv\in E} \frac{TP_{uv}}{n^X_{uv}}
+
p \operatorname*{mean}_{uv\in E} \frac{TP_{uv}}{n^X_{uv}}
\right).
\]

The design constraints restrict the set of feasible topologies, so not every tree corresponds to a valid network topology. For example, the master hub may be connected to at most 20 neighboring nodes, whereas every other node is limited to 11 neighbors. Any topology $T$ that violates one or more of these constraints is considered infeasible, and the objective value of the corresponding solution is penalized by setting $f(s(T))=-\infty$.

Despite its heuristic nature, the configuration procedure that produces solution $s(T)$ from topology $T$ remains computationally expensive. The key takeaway from this discussion is therefore that the number of times this procedure is invoked should be kept as small as possible.


\section{Literature review}\label{sec:literature_review}

Machine learning has attracted increasing attention in combinatorial optimization, particularly for problems where exact methods become computationally prohibitive on large-scale or complex instances. \citet{bengio2021machine} provide a comprehensive methodological overview of how learning techniques can be integrated into optimization algorithms, either to construct solutions directly or to support specific algorithmic decisions. More specifically, \citet{cappart2023combinatorial} investigate the use of graph neural networks for combinatorial optimization and reasoning, emphasizing their ability to exploit the underlying graph structure of many optimization problems. Similarly, \citet{peng2021graph} review graph learning approaches and discuss how graph-based representations can be leveraged to learn decision-making policies for structured optimization problems.


Several learning-based approaches have been proposed to enhance exact and heuristic optimization algorithms. For instance, \citet{khalil2016learning} introduced a machine learning framework for variable branching in mixed-integer programming, where the model learns branching decisions from strong branching information. \citet{gasse2019exact} represented mixed-integer programs as bipartite graphs and employed graph convolutional neural networks combined with imitation learning to improve branching decisions within branch-and-bound algorithms. Similarly, \citet{paulus2023learning} developed a learning-based diving strategy for branch-and-bound, in which graph neural networks predict variable assignments and guide primal heuristic decisions. These studies demonstrate that machine learning can effectively learn algorithmic decisions that are traditionally based on computationally expensive procedures or manually designed heuristics.

Many combinatorial optimization problems can be naturally represented as graphs, making graph neural networks particularly well suited for learning decision rules in such settings. \citet{khalil2017learning} were among the first to learn heuristics for graph-based combinatorial optimization problems by combining reinforcement learning with graph embeddings. \citet{li2018combinatorial} integrated graph convolutional networks with guided tree search to solve combinatorial optimization problems on graphs, demonstrating how learned models can guide search procedures without fully replacing them. These studies are particularly relevant to our work, as wireless network topologies naturally admit graph representations, and local search moves can be characterized through edge modifications.

Another important research direction focuses on the integration of machine learning into metaheuristic algorithms. \citet{talbi2021machine} provides a comprehensive taxonomy of the different ways in which learning techniques can be incorporated into metaheuristics, including parameter control, solution evaluation, move selection, and search guidance.
\citet{song2020general} introduced a learning-based large neighborhood search framework for integer linear programs, in which a model predicts partitions of the integer variables to define smaller subproblems that can be solved more efficiently. Similarly, \citet{wu2021learning} developed a deep reinforcement learning approach for learning large neighborhood search policies in integer programming, where the learned policy acts as a destroy operator by selecting variables for reoptimization. More recently, \citet{huang2023searching} employed contrastive learning to explore large neighborhoods for integer linear programs. \citet{niroumandrad2024learning} presented a learning-based Tabu Search approach for a scheduling problem, using classification models to reduce the search space and the computational effort required to evaluate candidate moves.
These studies are particularly relevant to our work, as they demonstrate that machine learning can be embedded within metaheuristics to guide the selection of promising moves while preserving the fundamental structure of the underlying search procedure.

In the context of network and telecommunication optimization, many problems involve complex combinatorial decisions related to topology design, routing, capacity allocation, and resource assignment. \citet{ribeiro2007metaheuristics} discuss the application of metaheuristics to optimization problems in computer communications, highlighting their relevance for network design and management. More recently, machine learning and deep learning techniques have gained increasing attention for addressing wireless networking problems. For example, \citet{zhang2019deep} review deep learning applications in mobile and wireless networks, including neural network-based approaches for resource management, traffic prediction, mobility analysis, and network control. \citet{luong2019applications} focus on deep reinforcement learning for communications and networking, where learning agents are employed to make sequential decisions in tasks such as dynamic access, routing, offloading, and resource allocation. Similarly, \citet{chen2019artificial} provide a tutorial on artificial neural network-based machine learning for wireless networks, covering supervised learning, unsupervised learning, and reinforcement learning approaches. These studies highlight the growing importance of learning-based methods in communication systems and motivate the development of structured learning models for optimization problems in which topology and link interactions play a fundamental role.


Graph-based learning approaches have also attracted increasing attention for wireless network optimization. \citet{vesselinova2020learning} review learning methods for graph-based combinatorial optimization, including applications to networking, and highlight the potential of graph machine learning models for solving network optimization problems. \citet{shen2020graph} proposed message-passing graph neural networks for scalable radio resource management, demonstrating that wireless resource allocation problems can be naturally formulated as graph optimization problems. \citet{eisen2020optimal} introduced random edge graph neural networks for wireless resource allocation and employed an unsupervised primal-dual learning framework to learn allocation policies over interference graphs. More recently, \citet{shen2022graph} examined the role of graph neural networks in wireless communications from both theoretical and practical perspectives, emphasizing their scalability and generalization capabilities in wireless network applications.

Closest to our work, \citet{feng} proposed a general machine learning framework for neighborhood generation in metaheuristic search and evaluated it on two combinatorial optimization problems. One of these applications is directly related to the wireless network optimization problem considered in this paper. Specifically, they integrated their framework into a Tabu Search algorithm for the tactical wireless network design problem introduced in \citet{2}, where candidate topologies are improved through edge-swap moves. In this setting, graph neural network classifiers are trained to identify promising edges to remove and add, thereby generating smaller yet more promising neighborhoods. This work is therefore the closest methodological reference to ours, as it demonstrates how supervised learning can effectively guide neighborhood generation within a Tabu Search framework for tactical wireless network design.

Our work follows the line of research initiated by \citet{feng} by integrating a drop-edge classifier into a Tabu Search procedure for tactical wireless network design. Unlike end-to-end learning approaches, the proposed method does not replace the optimization algorithm. Instead, the classifier is used to reduce and prioritize the set of candidate edges for removal during neighborhood generation, while feasibility checks, tabu restrictions, aspiration criteria, and objective evaluation remain fully handled by the underlying Tabu Search framework.

\section{A Tabu Search algorithm}\label{sec:topology_tabu_search}

Two tabu Search algorithms have been proposed for this network design problem \cite{WH1,2}. The algorithm presented in \cite{WH1} exhibits the best computational performance and is therefore adopted in this work. Its main components are outlined below, while a detailed description is available in the original publication.

\begin{figure}[!htb]
\centering
\begin{tikzpicture}[
    scale=1.05,
    every node/.style={circle, fill=black, inner sep=2pt},
    edge/.style={black!65, line width=1.2pt},
    dropedge/.style={blue, line width=1.6pt},
    addedge/.style={red, line width=1.6pt}
]

\begin{scope}[xshift=0cm]

    \node(A)  at (-0.5,0) {};
    \node(B2) at (0.5,0) {};
    \node(B)  at (-1,-0.5) {};  
    \node(C)  at (1,-0.5) {}; 
    \node(D)  at (-1.5,-1.25) {};
    \node(E)  at (-1,0.5) {};  
    \node(E2) at (-1.75,1.25) {}; 
    \node(E3) at (-2,0.5) {};
    \node(E4) at (-1,1.5) {};
    \node(G)  at (-1,-1.5) {};
    \node(F)  at (1,0.5) {};     
    \node(M)  at (-0.5,-1.25) {};
    \node(N)  at (0.25,-1.5) {};
    
    \draw[edge] (A) -- (B);
    \draw[edge] (E2) -- (E);
    \draw[edge] (E3) -- (E);
    \draw[edge] (E4) -- (E);
    \draw[dropedge] (A) -- (B2); 
    \draw[edge] (B2) -- (C);
    \draw[edge] (B) -- (D);
    \draw[edge] (B) -- (M);
    \draw[edge] (N) -- (M);
    \draw[edge] (B) -- (G);
    \draw[edge] (B2) -- (F);
    \draw[edge] (A) -- (E);

    \node[draw=none, fill=none, font=\small] at (-0.4,-2) {Current topology};
\end{scope}

\node[draw=none, fill=none, font=\Large] at (3.2,0) {$\Longrightarrow$};

\begin{scope}[xshift=7cm]

    \node(A2)  at (-0.5,0) {};
    \node(B22) at (0.5,0) {};
    \node(Bb)  at (-1,-0.5) {};  
    \node(C2)  at (1,-0.5) {}; 
    \node(D2)  at (-1.5,-1.25) {};
    \node(Ee)  at (-1,0.5) {};  
    \node(E22) at (-1.75,1.25) {}; 
    \node(E32) at (-2,0.5) {};
    \node(E42) at (-1,1.5) {};
    \node(G2)  at (-1,-1.5) {};
    \node(F2)  at (1,0.5) {};     
    \node(M2)  at (-0.5,-1.25) {};
    \node(N2)  at (0.25,-1.5) {};
    
    \draw[edge] (A2) -- (Bb);
    \draw[edge] (E22) -- (Ee);
    \draw[edge] (E32) -- (Ee);
    \draw[edge] (E42) -- (Ee);
    \draw[edge] (B22) -- (C2);
    \draw[edge] (Bb) -- (D2);
    \draw[edge] (Bb) -- (M2);
    \draw[edge] (N2) -- (M2);
    \draw[edge] (Bb) -- (G2);
    \draw[edge] (B22) -- (F2);
    \draw[edge] (A2) -- (Ee);

    \draw[addedge] (M2) -- (C2); 

    \node[draw=none, fill=none, font=\small] at (-0.4,-2) {Neighbor topology};

\end{scope}
\end{tikzpicture}
\vspace{-1cm}
\caption{Example of an edge-exchange move.}
\label{fig:edge_swap_example}
\end{figure}
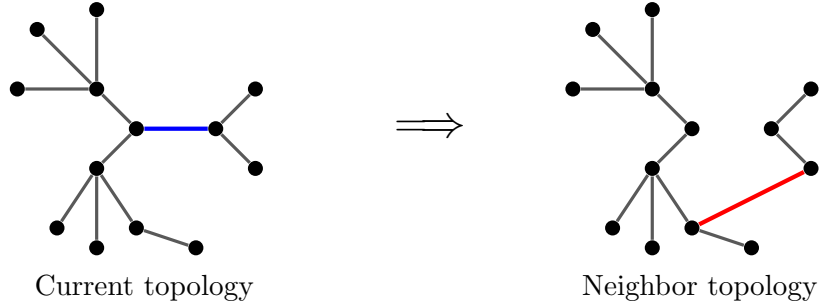

The algorithm starts from an initial feasible tree topology and iteratively explores neighboring topologies.
The neighborhood of a topology is defined by edge-exchange moves. More precisely, given a current tree topology $T=(V,E)$, an edge $e \in E$ is removed, splitting the tree into two connected components. A feasible edge $e' \notin E$ is then added to reconnect these components, yielding a new tree topology $T'$. This move preserves the tree structure while modifying the network communication paths.
This neighbor-generation process is illustrated in Figure~\ref{fig:edge_swap_example}, where the blue edge is replaced by the red edge.

Two tabu lists are maintained to avoid cycling. The list $L_{\mathrm{drop}}$ forbids the removal of recently added edges, whereas $L_{\mathrm{add}}$ forbids the reinsertion of recently removed edges. A move is admissible if it is not tabu or if it satisfies the aspiration criterion by producing a solution better than the best one found so far.

At each iteration, candidate edge-exchange moves are generated and evaluated, and the best admissible neighbor is selected as the next current solution. The tabu lists are then updated according to the performed move. This process is repeated until a stopping criterion is reached.

As discussed in Section \ref{sec:problem_description}, evaluating a topology $T$ requires applying the computationally expensive configuration procedure that constructs the corresponding solution $s(T)$. Consequently, the main computational bottleneck of the Tabu Search algorithm is the evaluation of neighboring topologies. For each candidate tree $T$, the configuration procedure must be executed to obtain $s(T)$ before its objective value can be computed. Since this procedure explores a large space of interdependent configuration decisions, it is computationally demanding, allowing only a limited number of Tabu Search iterations. The objective of this work is therefore to avoid evaluating every neighboring topology by exploiting machine learning techniques.

\section{A Machine learning approach for neighborhood reduction}\label{sec:tabu_ml}

We first present the overall idea behind the proposed machine learning approach, whose objective is to reduce the number of neighboring topologies that must be evaluated during the execution of the Tabu Search algorithm. The main motivation is to limit the computational effort associated with the exhaustive exploration of the neighborhood while preserving the ability of the search procedure to identify high-quality solutions.

We then describe in detail the learning process used to achieve this objective. In particular, we explain how the learning data are generated from previous Tabu Search trajectories, how the relevant features are defined, and how the classifier is trained to distinguish promising candidate moves from less promising ones.

Finally, we explain how the information provided by the learning phase is incorporated into the Tabu Search algorithm. More specifically, we describe how the predictions of the learned model are used to guide the selection of candidate moves and to focus the search on the most promising regions of the solution space.

\subsection{Overview of the proposed approach}\label{sec:learning_guided_move_generation}

As mentioned in the previous section, the main factor limiting the effectiveness of the Tabu Search algorithm described in \cite{WH1} is the relatively small number of iterations that can be carried out within a reasonable computational time. This limitation is mainly due to the fact that, at each iteration, a large number of neighboring topologies must be evaluated. Moreover, the evaluation of each candidate topology is itself computationally demanding, as it involves determining the corresponding network configuration and computing the associated objective function value. Consequently, only a limited number of iterations can be performed before the computational cost becomes prohibitive.

For a network composed of $n$ nodes, the number of neighboring topologies that must be considered at each iteration is of order $O(n^3)$. Indeed, the current tree topology contains $n-1$ edges that can potentially be removed, and each removed edge can be replaced by one among $O(n^2)$ possible edges in order to reconnect the two resulting components. This large neighborhood size is therefore a major contributor to the computational burden of the Tabu Search procedure.

A simple strategy to reduce the number of moves to be evaluated would consist in randomly sampling a subset of candidate moves. However, such a strategy does not take advantage of the information gathered during previous search iterations and may overlook promising topological transformations. Motivated by the learning-based approach introduced in \cite{feng}, we exploit data collected from previous Tabu Search trajectories to build a predictive model capable of identifying candidate moves that are more likely to lead to high-quality solutions during the search process.

More precisely, in order to reduce the number of neighboring topologies that need to be evaluated, we employ a machine learning technique to identify, among the $n-1$ edges of the current topology, those that appear to be the most promising candidates for removal. The objective is to determine which edge removals are most likely to lead, through an optimal exchange operation, to high-quality neighboring topologies, thereby avoiding the exhaustive evaluation of all possible edge-exchange moves.

The learning task is formulated as a supervised classification problem. Each candidate edge removal is characterized by a set of features capturing information about the current solution, the underlying graph structure, and the local impact of the considered modification. The classifier is trained using data collected from previous Tabu Search trajectories, with the objective of distinguishing promising candidates from less relevant ones.


\subsection{Training data generation}\label{sec:training_data_generation}

The classifier is trained using data generated from executions of the baseline Tabu Search algorithm presented in \cite{WH1}. During these executions, the algorithm iteratively explores a sequence of feasible tree topologies $T$ and their associated solutions $s(T)$. Each visited solution is considered as a training instance, for which we aim to identify the edge removals that can be involved in an edge-exchange move capable of improving the objective value compared with the current solution.
The data generation procedure used to train the classifier is now described in more detail. 

Let $s$ be one of these training instances, and let $T$ be its underlying topology. For each edge $e$ in $s$, we consider the possible edge-exchange moves obtained by removing $e$ from the topology. Let $\mathcal{A}(T,e)$ denote the set of feasible edges that can be introduced after the removal of $e$ to reconnect the two resulting components and restore a valid topology. Based on the potential of these edge-exchange moves to improve the current solution, a binary label is assigned to each edge $e$ in $s$, according to the following definition:

\[
y(s,e) =
\begin{cases}
1 & \text{if } \exists e' \in \mathcal{A}(T,e) \text{ such that } f(s(T-e+e'))>f(s),\\
0 & \text{otherwise}.
\end{cases}
\]
where $f(s(T-e+e'))$ denotes the objective value of the solution associated with the neighboring topology obtained by removing edge $e$ from $T$ and adding edge $e'$. Let $D(s)$ denote the subset of edges in solution $s$ such that $y(s,e)=1$. These are the edges whose removal leads, after applying the configuration procedure, to a neighboring topology associated with a solution having a higher objective value.
Let $N$ denote the number of training examples. The training dataset is defined as
\[
\mathcal{D} =
\left\{
\left(s_i,D(s_i)\right)
\right\}_{i=1}^{N},
\]
where each $s_i$ corresponds to a solution visited during an execution of the Tabu Search algorithm. Therefore, each training instance consists of a visited solution and the associated set of edges whose modification generates neighboring topologies that can be transformed into improving solutions.
In other words, the training dataset records, for each solution encountered during the Tabu Search executions, which edges represent promising moves because their removal leads to neighboring topologies whose associated solutions have a higher objective value. Edges in $s_i$ but not in $D(s_i)$ are considered non-promising, as their removal does not lead to an improving solution.



\subsection{A Graph Neural Network training model}\label{sec:GNN}

Since the tree topologies considered during the Tabu Search procedure are naturally represented as graph-structured objects, we adopt a graph neural network (GNN) to implement the drop-edge classifier. This choice follows the general learning-based methodology introduced in \cite{Scarselli}, where the objective is to exploit both the structural information of the graph representation and the descriptive information contained in the associated features. In our case, each solution is represented as a communication network graph, in which nodes and edges are described by dedicated sets of features. These features capture physical, structural, configuration, and performance-related information. By combining the attributes of individual nodes and edges with the underlying graph structure, the classifier can assess not only the intrinsic characteristics of a candidate edge but also its position and influence within the overall network.

Given a solution $s$, a set of descriptive features is extracted from both the nodes and the edges in order to capture the structural, spatial, and communication characteristics of the network.
For each node, the following features are considered:
\begin{itemize}[topsep=0pt, itemsep=0pt, parsep=0pt, partopsep=0pt, after=\vspace{3pt}]
\item its spatial coordinates, which describe its physical location within the deployment area;
\item its number of descendants in the rooted tree representation, providing information about the size of the subtree rooted at the node and its relative importance in the network hierarchy.
\end{itemize}

For each edge $uv$ of the tree topology, the following features are extracted to characterize its  contribution to the overall network performance:
\begin{itemize}[topsep=0pt, itemsep=0pt, parsep=0pt, partopsep=0pt, after=\vspace{3pt}]
\item the two end nodes defining the edge;
\item its path loss, reflecting the signal attenuation along the corresponding wireless link;
\item its fade margin, which measures the robustness of the communication link with respect to channel fluctuations;
\item its communication mode (PTP or PMP), indicating whether the link operates in a point-to-point or point-to-multipoint configuration;
\item its assigned communication channel;
\item its direct throughput $TP_{uv}$, corresponding to the nominal transmission capacity of the link;
\item its effective throughput $\frac{TP_{uv}}{n^X_{uv}}$ under each scenario $X\in\{A,B,C\}$, accounting for the sharing of the link capacity among the users traversing it;
\item its weighted throughput
$
\sum_{X\in\{A,B,C\}}\omega_X\frac{TP_{uv}}{n^X_{uv}},
$
which aggregates the effective throughput over all considered scenarios according to their respective weights.
\end{itemize}

Finally, the objective value $f(s)$ of solution $s$ is included as a global feature. It provides a quantitative measure of the solution quality according to the optimization criterion.

The chosen GNN architecture is similar to the one proposed in \cite{feng}. It consists of three main modules: an input module, a convolution module, and an output module. The input module encodes the features associated with the nodes and edges of the graph, while the convolution module progressively extracts hidden representations by propagating and aggregating information across neighboring nodes and edges. Finally, the output module maps the learned embedding of each edge in the current solution to a two-dimensional output vector.

Consequently, for each edge in the current solution, the GNN outputs a pair of numerical values, commonly referred to as \emph{logits}. These logits correspond to the raw prediction scores associated with the two possible classes, namely the \textit{non-improving} and \textit{improving} classes, before the application of any normalization function.

To obtain interpretable predictions, the two logits associated with each edge are passed through a softmax activation function. This operation transforms the raw scores into two probabilities whose sum is equal to one, thereby providing a probability distribution over the two classes. The resulting probabilities quantify the confidence of the GNN in assigning the edge to either class and are subsequently used to guide the edge selection process during the optimization procedure.

\subsection{Learning the GNN parameters}\label{sec:Learning}
The objective of the learning phase is to determine a parameter configuration $\theta$ of the GNN that is able to accurately distinguish between promising and non-promising candidate edges for the drop operation. More specifically, given the training dataset $\mathcal{D}$, the learning algorithm seeks to adjust the network parameters so that the predictions produced by the GNN are consistent with the labels observed in the training samples.

For each training example $(s_i,D(s_i)) \in \mathcal{D}$, the GNN receives the solution $s_i$ together with the corresponding node and edge features as input. The objective of the classifier is to predict which edges in $s_i$ are promising candidates for removal from the current topology. More specifically, the classifier is expected to assign edge $e$ in $s_i$ to the \textit{improving} class if $e \in D(s_i)$, meaning that its removal can lead to an improving edge-exchange move. Otherwise, the edge should be classified as belonging to the \textit{non-improving} class.

This binary classification problem is inherently affected by class imbalance. Indeed, among all removable edges of a topology, only a relatively small fraction is expected to produce an improving move when removed, whereas the vast majority corresponds to non-improving candidates. As a consequence, a standard cross-entropy loss would tend to bias the learning process toward the majority class. To alleviate this issue, the GNN is trained using a weighted cross-entropy loss, where a larger weight is assigned to the improving class. This weighting strategy penalizes misclassification of improving edges more heavily and therefore encourages the model to identify promising drop candidates despite their limited number.

More precisely, let $\hat{y}^\theta(s,e)$ denote the probability predicted by the GNN, with parameter configuration $\theta$, that the candidate edge $e$ in solution $s$ belongs to the improving class. For a given training sample $(s_i,D(s_i))$, the prediction error $\mathcal{L}(\theta,i)$ is measured using the following weighted cross-entropy loss, where $E(s_i)$ denotes the set of edges in $s_i$:
\[
\mathcal{L}(\theta,i)
=
-
\left(
\lambda\sum_{e\in D(s_i)}
 \log\!\left(\hat{y}^\theta(s_i,e)\right)
\;\;+\;\;
(1-\lambda)
\sum_{e\in E(s_i)\setminus D(s_i)}\log\!\left(1-\hat{y}^\theta(s_i,e)\right)
\right),
\]
where the parameter $\lambda\in[0,1]$ controls the relative importance assigned to the improving class with respect to the non-improving class. Increasing the value of $\lambda$ places greater emphasis on correctly identifying improving edges during the training process.

The optimal parameter configuration of the GNN is finally obtained by minimizing the cumulative loss over the entire training dataset. Formally, the learning problem can be expressed as the following supervised optimization problem:
\[
\theta^{*}
=
\arg\min_{\theta}
\sum_{i=1}^{|\mathcal{D}|}
\mathcal{L}(\theta,i).
\]

The resulting parameter configuration $\theta^{*}$ corresponds to the model that best fits the training data according to the weighted loss function, while accounting for the imbalance between improving and non-improving examples.

For this purpose, the available dataset is randomly partitioned into two disjoint subsets. The first subset, containing 80\% of the samples, is used to train the parameters of the GNN by minimizing the weighted cross-entropy loss over the training examples. The remaining 20\% of the samples form the validation set and are not used during the parameter update process.

After each training phase, the performance of the learned model is evaluated on the validation set. This evaluation provides an estimate of the model's ability to generalize to unseen data and is used to compare different values of the hyperparameter $\lambda$. Since $\lambda$ determines the relative importance assigned to the improving class in the weighted loss function, its value has a direct impact on the trade-off between correctly identifying promising edges and limiting false positive predictions. The value of $\lambda$ retained for the final model is therefore the one leading to the best predictive performance on the validation set.

\subsection{ML-guided Tabu Search}\label{sec:ml_guided_tabu}

The trained drop-edge classifier is integrated into the Tabu Search procedure to guide the generation of edge-exchange neighbors. Instead of exhaustively exploring the entire neighborhood at each iteration, the classifier is used to evaluate the removable edges of the current solution and to estimate their potential to produce improving neighboring topologies. For each candidate edge, the GNN outputs a probability representing the likelihood that removing this edge will lead to an improving move.

The proposed approach does not use the classifier as a strict binary decision mechanism that simply accepts or rejects candidate edges. Such a strategy could excessively restrict the search space and even lead to an empty or poorly diversified neighborhood, especially when only a few edges are predicted to be promising or when some of these edges are temporarily forbidden by the tabu restrictions. Instead, all candidate removable edges are ranked according to their predicted probabilities, from the most to the least promising. A fixed proportion $\tau$ of the highest-ranked edges is then retained for neighborhood generation.

This ranking-based strategy substantially reduces the number of edge-exchange moves that must be evaluated while preserving the exploratory capability of the search. Indeed, because several high-scoring candidate edges are retained rather than only those classified as improving, the resulting neighborhood remains sufficiently diversified after feasibility constraints and tabu restrictions have been applied. This allows the search to focus on the most promising regions of the solution space without sacrificing its ability to escape local optima.

For each selected edge, the algorithm removes the edge from the current topology and identifies the two connected components induced by this removal. The set of feasible reconnection edges linking these two components is then generated, and the corresponding edge-exchange neighbors are evaluated. Among all admissible neighboring solutions, the next solution is selected according to the standard Tabu Search selection rules.

\begin{figure}[H]
    \centering
    \includegraphics[width=0.85\linewidth]{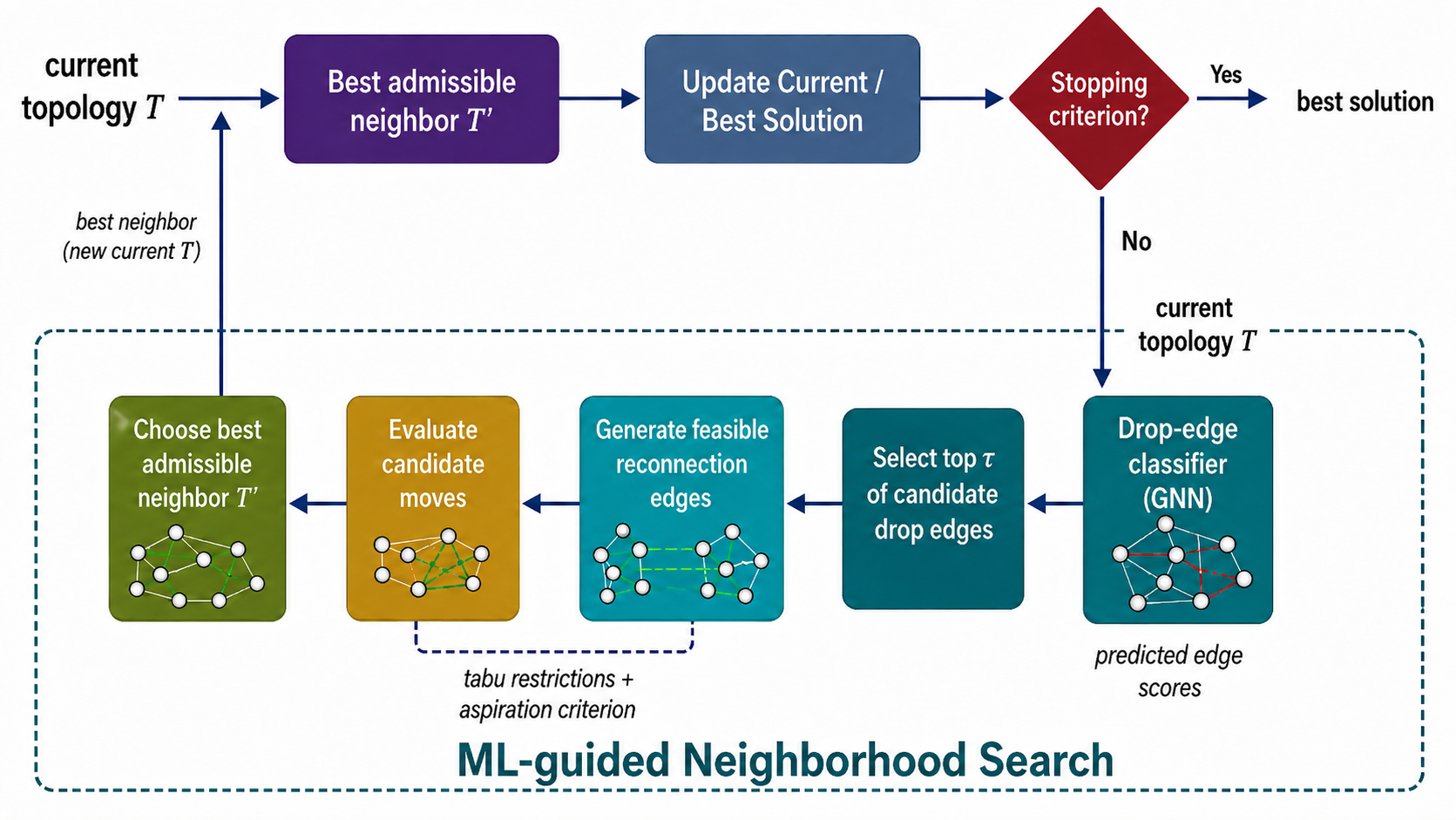}
\caption{Schematic representation of the ML-guided neighborhood search.}
\label{fig:ml_guided_neighborhood}

    \label{fig:placeholder}
\end{figure}

The integration of the GNN classifier preserves the overall structure of the baseline Tabu Search algorithm. The machine learning component intervenes exclusively during the neighborhood generation phase by prioritizing the candidate edges considered for removal. All other aspects of the search procedure remain unchanged, including the generation of feasible reconnection edges, the evaluation of neighboring solutions, the management of tabu restrictions and the aspiration criterion, as well as the updates of the current and best-known solutions. Consequently, the proposed approach can be viewed as a lightweight enhancement of the original Tabu Search, where the GNN focuses the search on the most promising edge removals without modifying the underlying optimization mechanism.
Figure~\ref{fig:ml_guided_neighborhood} illustrates the overall ML-guided Tabu Search framework.

As previously mentioned, evaluating a topology $T$ requires completing several configuration steps. More specifically, the master hub and the partition of its neighboring nodes must first be determined. Communication channels and operating frequencies are then assigned, and the set of active antenna beams together with their corresponding orientations is selected. 

As observed in \cite{WH1}, if a node is identified as a good choice for the master hub in a topology $T$, then it is likely to remain a suitable master hub for a neighboring topology $T'$, obtained from $T$ by replacing a single edge.
Likewise, the partition of the neighbors of the master hub is expected to remain largely unchanged between $T$ and $T'$, requiring only minor adjustments to account for the insertion or deletion of a single edge. More precisely, consider a topology $T$ with master hub $r$ and a partition $\pi$ of its neighbors, and let $T'$ be a neighboring topology obtained by replacing an edge $e$ of $T$ with another edge $e'$. If the removed edge $e$ is incident to $r$ and one of its neighbors $u$, then $u$ is removed from its corresponding block in the partition. Conversely, if the added edge $e'$ is incident to $r$ and a vertex $v$, then $v$ is inserted into the block containing the neighbor $w$ of $r$ whose incident angle with $r$ is closest to that of $v$.

The Tabu Search algorithm proposed in \cite{WH1} exploits these observations by modifying the configuration procedure that constructs solution $s(T)$ from topology $T$. It preserves the same master hub and the partition of its neighbors, applying only the minor adjustments described above when necessary. The original configuration procedure, which reconsiders both the master hub and the partition of its neighbors, is invoked only once every $n$ iterations, where $n$ is the number of nodes.

As mentioned in the previous section, $\hat{y}^{\theta^*}(s,e)$ denotes the probability assigned by the GNN with the chosen parameter configuration $\theta^*$ to the improving class for edge $e$ in solution $s$. The ML-guided Tabu Search algorithm is outlined below.

\begin{algorithm}[H]
\setstretch{0.98}
\renewcommand\thealgorithm{}
\caption{\texttt{TABU-ML}\\
{\bf Input}: a set of $n$ nodes, a parameter $\tau$, and a trained GNN that outputs a score $\hat{y}^{\theta^*}(s,e)$ for each edge $e$ of a solution $s$.\\
{\bf Output}: a solution $s^*$.}
\label{alg:MLTABU}
\begin{algorithmic}[1]
    \STATE Generate an initial topology $T$ as described in \cite{WH1}, determine $s(T)$ and set $s\gets s(T)$.
    \STATE Set $T^*\gets T$, $s^*\gets s(T)$, $f^*\gets f(s)$, $L_{drop}\gets\emptyset$, and $L_{add}\gets\emptyset$.
    \WHILE{the time limit is not reached}
        \STATE Compute the values $\hat{y}^{\theta^*}(s,e)$ predicted by the GNN for each edge $e$ in $s$, and construct a list $L$ containing the $\lceil \tau(n-1)\rceil$ edges with the highest values. Set $f_{best}\gets-\infty$.
        \FOR{each edge $e$ in $L$}
            \STATE Remove $e$ from $T$ and let $C_1$ and $C_2$ be the two resulting connected components.
            \FOR{each edge $e'\neq e$ reconnecting $C_1$ and $C_2$}
                \STATE Let $T'$ be obtained from $T$ by replacing $e$ by $e'$.
                \STATE Determine $s(T')$ by retaining the master hub and neighbor partition of $T$, except every $n$-th iteration, when both are reconsidered; set $s'\gets s(T')$ and $f'\gets f(s')$.
                \IF{($f'>f^*$) or ($f'>f_{best}$, $e\notin L_{drop}$ and $e'\notin L_{add})$}
                    \STATE Set  $T_{best}\gets T'$, $s_{best}\gets s'$ and $f_{best}\gets f'$.
                \ENDIF
            \ENDFOR
        \ENDFOR
        \STATE Let $e^d$ and $e^a$ denote the edge removed from and added to $T$, respectively, to obtain $T_{best}$.
        \STATE Add $e^a$ to $L_{drop}$, add $e^d$ to $L_{add}$ and set $T\gets T_{best}$ and $s\gets s_{best}$.
        \IF{$f_{best}>f^*$}
            \STATE Set $T^*\gets T$, $s^*\gets s$ and $f^*\gets f_{best}$.
        \ENDIF
    \ENDWHILE
    \STATE \textbf{return} $s^*$.
\end{algorithmic}
\end{algorithm}

By setting $\tau=1$, the algorithm reduces to the original Tabu Search proposed in \cite{WH1}. The most computationally expensive part of the algorithm is Step 9, where $s(T')$ must be determined for trees $T'$ obtained by replacing one edge of $T$ with another. This requires assigning communication channels and frequencies, determining the active antenna beams and their orientations, and, when necessary, selecting the master hub and partitioning its neighboring nodes.

In contrast, setting $\tau=0.2$, the value adopted in the computational experiments, significantly reduces the number of these costly evaluations by restricting them to the candidate trees predicted to produce the best objective values.

\section{Computational experiments}\label{sec:computational_experiments}

This section presents the computational experiments conducted to evaluate the performance of the proposed ML-guided Tabu Search algorithm. The main objective of these experiments is to determine whether the trained drop-edge classifier can effectively reduce the size of the explored neighborhood while preserving the solution quality achieved by the baseline Tabu Search algorithm, and potentially enable the search to reach better solutions within the same computational time. The experimental methodology is described in detail, including the generation of the test instances, the training procedure of the classifier, the performance evaluation protocol, and the computational environment used for the experiments. Finally, the computational results are presented and discussed.

\subsection{Experimental setting}\label{sec:experimental_setting}

\subsubsection{Instance generation}\label{sec:instance_generation}

The computational experiments are conducted on synthetic tactical wireless network instances generated according to the procedure described in \cite{WH1}. Each instance consists of a set of geographically distributed nodes together with the physical parameters required to evaluate radio links, including path losses and fade margins. These data are used to compute the signal quality, interference levels, and throughput associated with candidate network configurations.

To assess the performance of the proposed approach under different problem sizes, we consider instances with 20, 30, and 50 nodes. For each network size, several independently generated instances are used to account for the variability resulting from the random placement of the nodes and the corresponding radio-link characteristics. This experimental setting provides a representative evaluation of the algorithm across instances of varying complexity.

\subsubsection{Classifier training setup}\label{sec:classifier_training_setup}

As explained in Section \ref{sec:training_data_generation}, the classifier is trained using data collected from executions of the Tabu Search algorithm proposed in \cite{WH1}. More specifically, each training pair $(s_i,D(s_i))\in\mathcal{D}$ can be associated with  binary values $y(s_i,e)$ for every edge $e$ of solution $s_i$, such that $y(s_i,e)=1$ if and only if $e\in D(s_i)$. In other words, $y(s_i,e)=1$ indicates that removing edge $e$ and performing a valid edge exchange leads to a neighboring topology whose associated solution has a higher objective value than the current solution. Consequently, each solution $s_i$ is associated with $n-1$ binary values $y(s_i,e)$, where $n$ denotes the number of nodes in the network.

To generate the training data, the Tabu Search algorithm is executed for one hour on instances with $n=20$ nodes, two hours on instances with $n=30$ nodes, and five hours on instances with $n=50$ nodes. This yields the following datasets:
\begin{itemize}[topsep=0pt, itemsep=0pt, parsep=0pt, partopsep=0pt, after=\vspace{3pt}]
\item for $n=20$: 48,766 visited solutions, yielding 926,554 binary values $y(s_i,e)$;
\item for $n=30$: 16,947 visited solutions, yielding 491,463 binary values $y(s_i,e)$;
\item for $n=50$: 8,199 visited solutions, yielding 401,751 binary values $y(s_i,e)$.
\end{itemize}

For each pair $(s_i,D(s_i))$, the features characterizing the solution $s_i$ are extracted and stored, as described in Section \ref{sec:GNN}. These features provide a representation of the solution and its underlying topology, while the associated set $D(s_i)$ identifies the edges corresponding to improving moves. Together, the extracted features of $s_i$ and the corresponding set $D(s_i)$ form the samples of the dataset used to train and evaluate the classifier. The complete dataset is then randomly divided into two subsets: 80\% of the samples are assigned to the training set, while the remaining 20\% are reserved for validation. This split allows the model parameters to be learned from the training samples while using the validation set to monitor performance on previously unseen data.

The model is trained using the Adam (Adaptive Moment Estimation) optimizer \cite{kingma}, a widely used and computationally efficient gradient-based optimization algorithm for training deep learning models. The optimizer is configured with a learning rate of $10^{-4}$ and a batch size of one graph observation. The training process is carried out for a maximum of 200 epochs.

As discussed in Section \ref{sec:Learning}, the dataset is imbalanced, with edges corresponding to improving moves being significantly underrepresented. To address this issue, a weighted cross-entropy loss function is employed, assigning a larger weight to the improving class in order to penalize misclassification of these samples more strongly. In the experiments presented below, the weight assigned to the improving class is set to $\lambda=0.95$.


\subsubsection{Performance evaluation protocol}\label{sec:evaluation_protocol}

The proposed ML-guided Tabu Search is evaluated against two reference variants in order to assess the contribution of the learned drop-edge classifier. The three compared methods are defined as follows:

\begin{itemize}[topsep=0pt, itemsep=0pt, parsep=0pt, partopsep=0pt, after=\vspace{3pt}]
\item \texttt{TABU}: the baseline Tabu Search algorithm proposed in \cite{WH1}, which considers the entire set of admissible droppable edges at each iteration;

\item \texttt{TABU-20\%}: a reduced-neighborhood version of the baseline algorithm, which randomly selects and evaluates only 20\% of the admissible droppable edges at each iteration;

\item \texttt{TABU-ML}: the proposed ML-guided variant, which selects the 20\% of admissible droppable edges with the highest scores according to the trained classifier.

\end{itemize}

All compared methods share the same Tabu Search framework, including the same initialization procedure, objective evaluation, feasibility checks, aspiration criterion, stopping condition, and time limit. The only difference between them is the strategy used to construct the set of candidate droppable edges before generating the edge-exchange neighborhood.


The length of the tabu-drop list $L_{drop}$ is chosen to be of the order of the square root of the number of possible reconnection edges. It is defined as
$$
|L_{drop}|=\max\left(3,\left\lceil\sqrt{\tau(n-1)}\right\rceil\right),$$
where $\tau=1$ for \texttt{TABU} and $\tau=0.2$ for the two other algorithms. For all three algorithms, the length of the tabu-add list $L_{add}$ is chosen to be of the order of the square root of the number of possible reconnection edges. It is defined as
\[
|L_{add}|
=
\max
\left(3,\left\lceil\sqrt{\frac{n(n-1)}{2}}\right\rceil\right).
\]

\subsubsection{Computational environment}\label{sec:computational_environment}

The drop-edge classifier is trained on a computing server using Python 3.9.25, PyTorch~\cite{paszke2019pytorch}, PyTorch Geometric~\cite{fey2019fast}, and CUDA 12.8. The server runs AlmaLinux 9.8 and is equipped with NVIDIA A100 GPUs.
After training, the classifier is integrated into the ML-guided Tabu Search algorithm. The computational comparison between the baseline Tabu Search and the two variants is performed on a separate workstation equipped with an Intel\textsuperscript{\textregistered} Core\textsuperscript{TM} i7-12700 processor, 64 GB of RAM, and a 64-bit AlmaLinux operating system. All algorithms included in the comparison are executed on the CPU using a single thread.


\subsection{Computational results}\label{sec:Results}

For the evaluation, we generate five instances with $n=20$ nodes and five instances with $n=30$ nodes, as well as three instances with $n=50$ nodes. They are generated independently and are distinct from the instances used to train the classifier, ensuring that the evaluation is performed on previously unseen network configurations.

The algorithms are compared using the following performance metrics. For each instance, we analyze the quality of the solution obtained at the end of the allowed computation time, as well as the number of iterations performed during the search. Since the algorithms are stochastic, each instance is executed five times independently. For each performance measure, we report the best value, the worst value, and the average value observed over these five runs.

The allowed computation time is adjusted according to the size of the instances, with limits of one hour for 20-node instances, two hours for 30-node instances, and five hours for 50-node instances.
Table~\ref{tab:Table_TABU_ML} summarizes the performance results of the three algorithms, allowing a direct comparison of their solution quality and search behavior across the different test instances.

\begin{table}[H]
\setlength{\tabcolsep}{3pt}
\centering
\footnotesize
\caption{Performance comparison of \texttt{TABU}, \texttt{TABU-20\%}, and \texttt{TABU-ML}}
\label{tab:Table_TABU_ML}
\vspace{2mm}

\resizebox{1.0\textwidth}{!}{
\begin{tabular}{cc|rrr|rrr|rrr||rrr|rrr}
\multicolumn{2}{c}{} &
\multicolumn{9}{c||}{\textbf{solution value}} &
\multicolumn{6}{c}{\textbf{iterations}} \\

\multicolumn{2}{c}{} &
\multicolumn{3}{c|}{TABU} &
\multicolumn{3}{c|}{TABU-20\%} &
\multicolumn{3}{c||}{TABU-ML} &
\multicolumn{3}{c|}{TABU} &
\multicolumn{3}{c}{TABU-ML} \\

\multicolumn{1}{c}{$n$} &
\multicolumn{1}{c|}{instance} &
best & avg & worst &
best & avg & worst &
best & avg & worst &
best & avg & worst &
best & avg & worst \\

\hline
\multirow{5}{*}{20}
& 1 & 34.67 & 34.32 & 33.94 & 33.80 & 31.82 & 30.72 & 34.59 & 34.59 & 34.59 & 12601 & 12461 & 12934 & 70775 & 69797 & 67958 \\
& 2 & 32.84 & 32.27 & 31.51 & 31.70 & 30.32 & 28.59 & 34.60 & 33.91 & 32.82 & 10747 & 10905 & 11499 & 60683 & 61074 & 60478 \\
& 3 & 33.79 & 33.44 & 32.91 & 32.82 & 32.16 & 31.65 & 33.84 & 33.82 & 33.78 & 13339 & 13176 & 13230 & 69390 & 67856 & 68066 \\
& 4 & 32.46 & 32.34 & 32.27 & 32.42 & 30.60 & 29.58 & 32.46 & 32.46 & 32.46 & 13730 & 13530 & 13764 & 76931 & 74392 & 72534 \\
& 5 & 32.00 & 31.65 & 30.75 & 30.86 & 29.72 & 29.00 & 32.01 & 31.95 & 31.88 & 13092 & 12460 & 12251 & 70449 & 66072 & 60879 \\

\cline{2-17}
& average & 33.15 & 32.80 & 32.28 & 32.32 & 30.92 & 29.91 & 33.50 & 33.34 & 33.11 & 12702 & 12506 & 12736 & 69646 & 67838 & 65983 \\

\hline
\multirow{5}{*}{30}
& 1 & 26.93 & 26.64 & 25.71 & 25.62 & 24.87 & 23.88 & 28.68 & 27.59 & 26.92 & 4613 & 4523 & 4371 & 27667 & 27052 & 25667 \\
& 2 & 31.64 & 28.33 & 25.74 & 25.42 & 25.18 & 24.56 & 31.60 & 30.60 & 29.51 & 4094 & 4014 & 3622 & 26152 & 25408 & 24288 \\
& 3 & 26.75 & 26.19 & 25.66 & 25.05 & 24.64 & 23.81 & 30.55 & 28.30 & 26.54 & 3520 & 4014 & 4193 & 25711 & 26212 & 25134 \\
& 4 & 30.47 & 28.18 & 26.70 & 25.23 & 24.92 & 24.56 & 30.58 & 29.54 & 27.78 & 4611 & 4192 & 4074 & 26187 & 24592 & 23994 \\
& 5 & 28.65 & 26.39 & 24.76 & 25.14 & 24.46 & 23.47 & 27.32 & 26.78 & 26.06 & 4177 & 3793 & 3331 & 24778 & 25147 & 24426 \\

\cline{2-17}
& average & 28.89 & 27.15 & 25.71 & 25.29 & 24.81 & 24.06 & 29.75 & 28.56 & 27.36 & 4203 & 4107 & 3918 & 26100 & 25682 & 24702 \\

\hline

\multirow{5}{*}{50}
& 1 & 16.72 & 15.62 & 14.89 & 15.68 & 14.76 & 13.90 & 17.23 & 16.58 & 15.10 & 836 & 757 & 474 & 7474 & 7606 & 7760 \\
& 2 & 18.93 & 16.40 & 13.31 & 17.61 & 15.91 & 14.77 & 19.63 & 17.94 & 16.47 & 1047 & 839 & 416 & 8623 & 8097 & 7879 \\
& 3 & 17.61 & 16.32 & 14.91 & 17.13 & 16.12 & 14.94 & 19.51 & 17.37 & 16.09 & 556 & 701 & 645 & 8051 & 7881 & 8040 \\

\cline{2-17}
& average & 17.75 & 16.11 & 14.37 & 16.80 & 15.60 & 14.54 & 18.79 & 17.30 & 15.89 & 813 & 766 & 512 & 8049 & 7861 & 7893 \\
\hline
\end{tabular}
}
\end{table}

The inclusion of \texttt{TABU-20\%} provides an important comparison point in order to better understand the source of the improvements obtained with \texttt{TABU-ML}. This variant has the same reduced neighborhood size as \texttt{TABU-ML}, but the candidate drop edges are selected randomly rather than according to the predictions of the learned classifier. It therefore allows us to distinguish between the effect of reducing the number of evaluated moves and the effect of intelligently guiding the neighborhood exploration.

If the performance gains of \texttt{TABU-ML} were mainly due to the evaluation of a smaller number of candidate moves, then \texttt{TABU-20\%} should exhibit a similar behavior. However, the results show that this is not the case. For all considered graph sizes, \texttt{TABU-20\%} generally obtains lower solution values than both \texttt{TABU} and \texttt{TABU-ML}. This indicates that randomly removing 80\% of the candidate drop edges can discard promising moves and reduce the ability of the search process to explore favorable regions of the solution space. In contrast, the classifier-based selection used by \texttt{TABU-ML} preserves more relevant candidate moves, allowing the algorithm to maintain or improve solution quality while exploring a smaller neighborhood.

Notably, \texttt{TABU-ML} consistently outperforms \texttt{TABU-20\%} across all considered instance sizes. For the 20-node instances, the average solution value over all test instances increases from 30.92 with \texttt{TABU-20\%} to 33.34 with \texttt{TABU-ML}. Similarly, for the 30-node instances, the average value improves from 24.81 to 28.56, while for the 50-node instances it increases from 15.60 to 17.30. These results demonstrate that selecting candidate edges based on the predictions of the learned classifier is substantially more effective than randomly selecting the same proportion of candidate edges. The improvement confirms that the performance gain of \texttt{TABU-ML} does not come solely from reducing the neighborhood size, but from the ability of the classifier to identify more promising moves.

The comparison with the baseline \texttt{TABU} further demonstrates that \texttt{TABU-ML} improves the overall solution quality. For the 20-node instances, the average solution value increases from 32.80 with \texttt{TABU} to 33.34 with \texttt{TABU-ML}. Similarly, for the 30-node instances, the average value improves from 27.15 to 28.56, while for the 50-node instances it increases from 16.11 to 17.30. These results show that the learned classifier does not merely reduce the computational effort by limiting the number of evaluated moves. Instead, it effectively guides the search toward more promising regions of the solution space, enabling the algorithm to identify higher-quality solutions within the same time limit.

A similar trend is observed when analyzing the worst solution values obtained over the five independent runs. For the 20-node instances, the average worst solution value increases from 32.28 with \texttt{TABU} and 29.91 with \texttt{TABU-20\%} to 33.11 with \texttt{TABU-ML}. For the 30-node instances, it improves from 25.71 with \texttt{TABU} and 24.06 with \texttt{TABU-20\%} to 27.36 with \texttt{TABU-ML}. Similarly, for the 50-node instances, the average worst value increases from 14.37 with \texttt{TABU} and 14.54 with \texttt{TABU-20\%} to 15.89 with \texttt{TABU-ML}. These results indicate that \texttt{TABU-ML} is not only able to achieve better average performance but also provides more consistent results across independent executions. By maintaining higher worst-case solution values, the ML-guided approach demonstrates greater robustness and a better ability to avoid poor-quality search trajectories.

However, \texttt{TABU-ML} does not outperform \texttt{TABU} on every individual instance. In particular, for some of the 20-node and 30-node instances, the best solution obtained by the baseline \texttt{TABU} algorithm is slightly better than the best solution found by \texttt{TABU-ML}. This behavior is expected, as the reduction of the neighborhood size may occasionally exclude specific promising moves that would have been explored by the full-neighborhood search. Nevertheless, when considering the results aggregated over all test instances, \texttt{TABU-ML} consistently demonstrates better overall performance, with higher average and worst solution values. These results indicate that the classifier-guided selection of candidate edges provides a more effective search strategy on average, even though it does not guarantee an improvement for every individual instance.

Regarding the number of iterations performed within the time limit, \texttt{TABU-ML} is able to execute more iterations than the baseline \texttt{TABU}. This improvement is a direct consequence of evaluating a smaller subset of candidate moves at each iteration, which reduces the computational effort required to explore each neighborhood. However, the results obtained with \texttt{TABU-20\%} show that simply increasing the number of iterations is not sufficient to improve the overall performance of the search. Although this variant benefits from the same reduction in neighborhood size and therefore performs more iterations, its solution quality remains lower than that of \texttt{TABU-ML}.

These observations highlight the importance of the quality of the selected candidate moves, rather than only the number of explored neighborhoods. By using the learned classifier to identify and retain more promising drop edges, \texttt{TABU-ML} combines the computational advantage of a reduced neighborhood with a more effective move-selection strategy. This combination allows the algorithm to explore the solution space more efficiently and explains why it achieves better solution values than both the full-neighborhood baseline and the randomly reduced neighborhood variant.

\begin{figure}[H]
\centering

\begin{subfigure}[b]{0.32\textwidth}
    \centering
    \includegraphics[width=\textwidth]{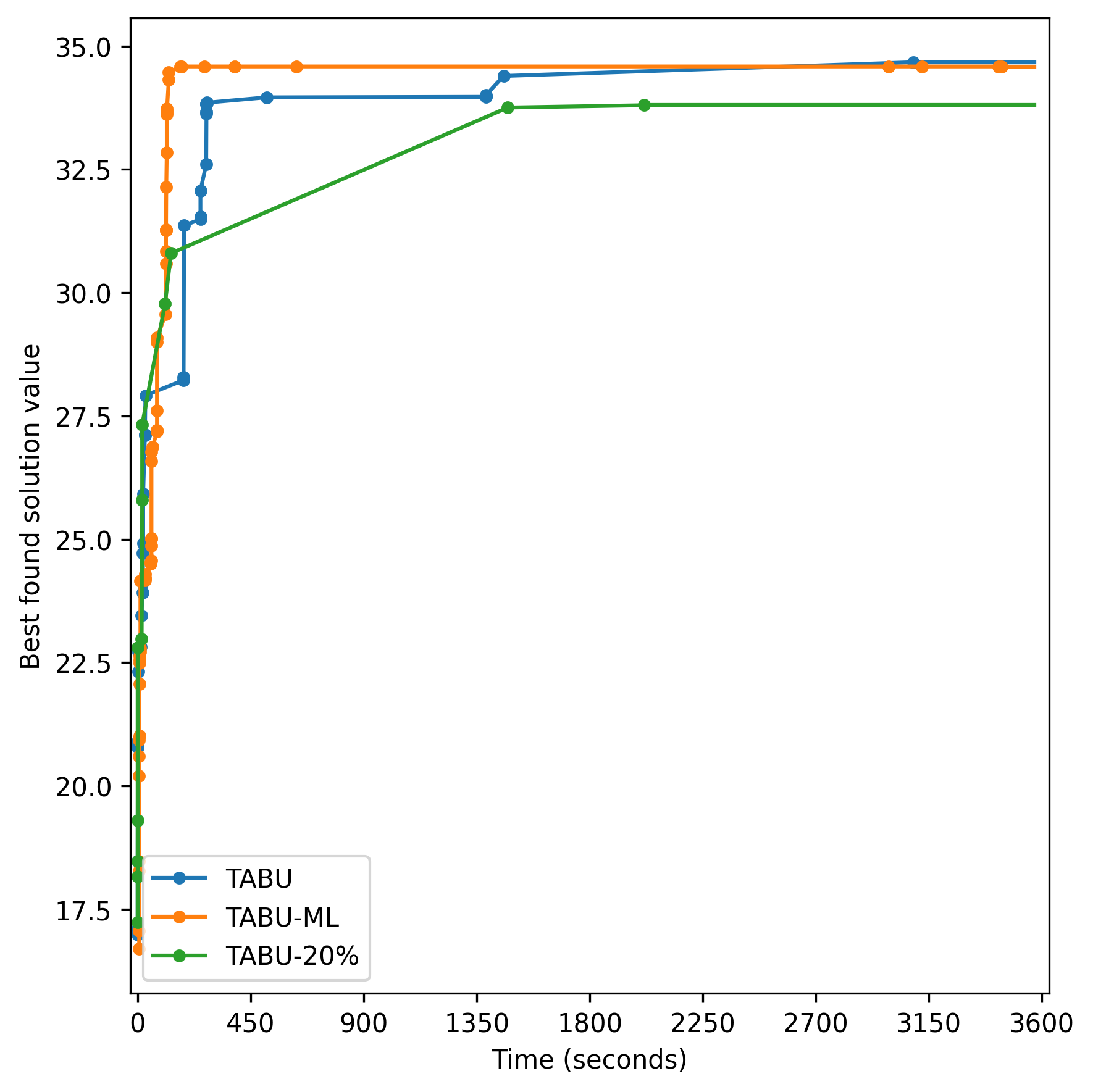}
    \caption*{20 nodes}
\end{subfigure}
\hfill
\begin{subfigure}[b]{0.32\textwidth}
    \centering
    \includegraphics[width=\textwidth]{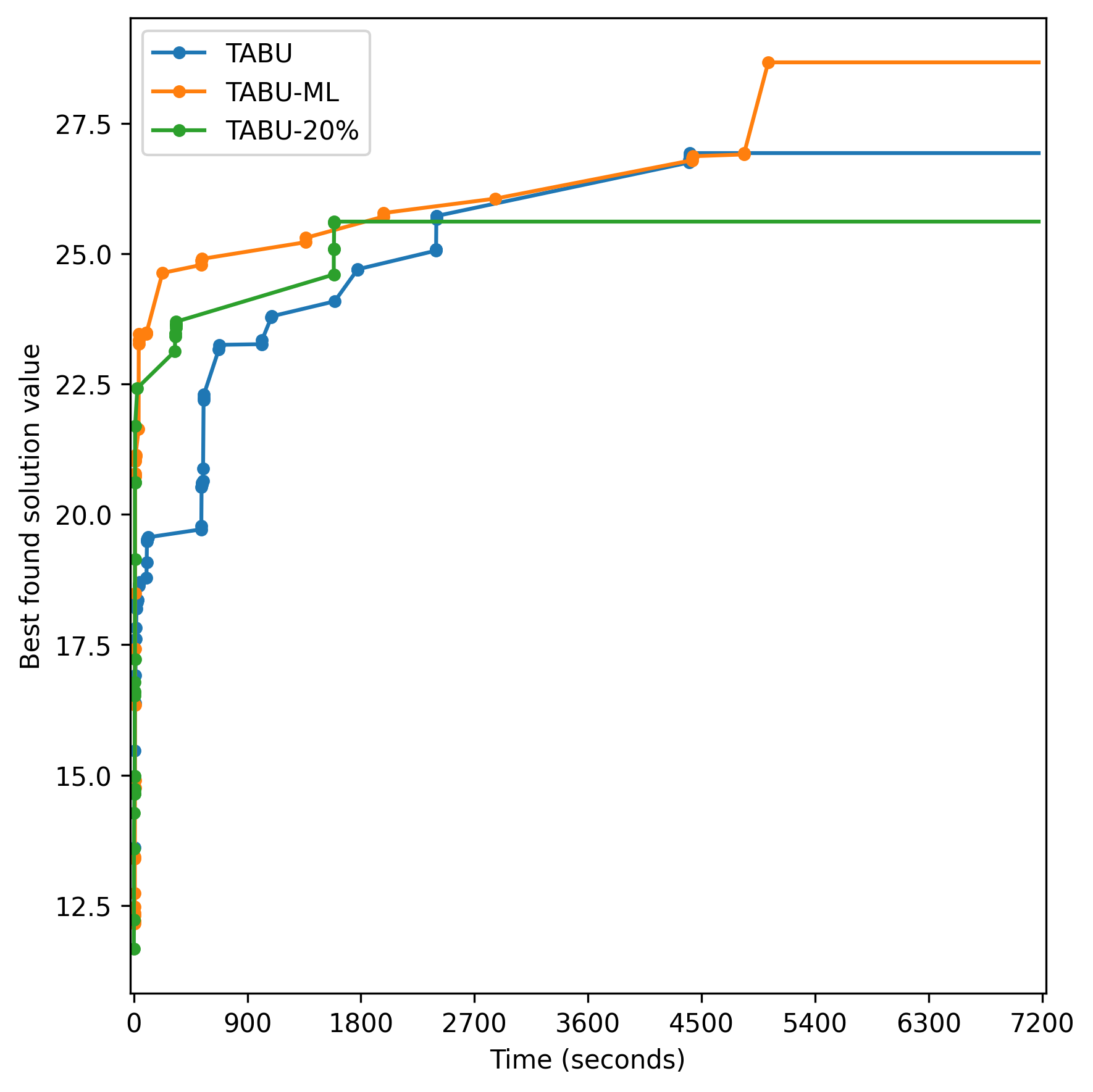}
    \caption*{30 nodes}
\end{subfigure}
\hfill
\begin{subfigure}[b]{0.32\textwidth}
    \centering
    \includegraphics[width=\textwidth]{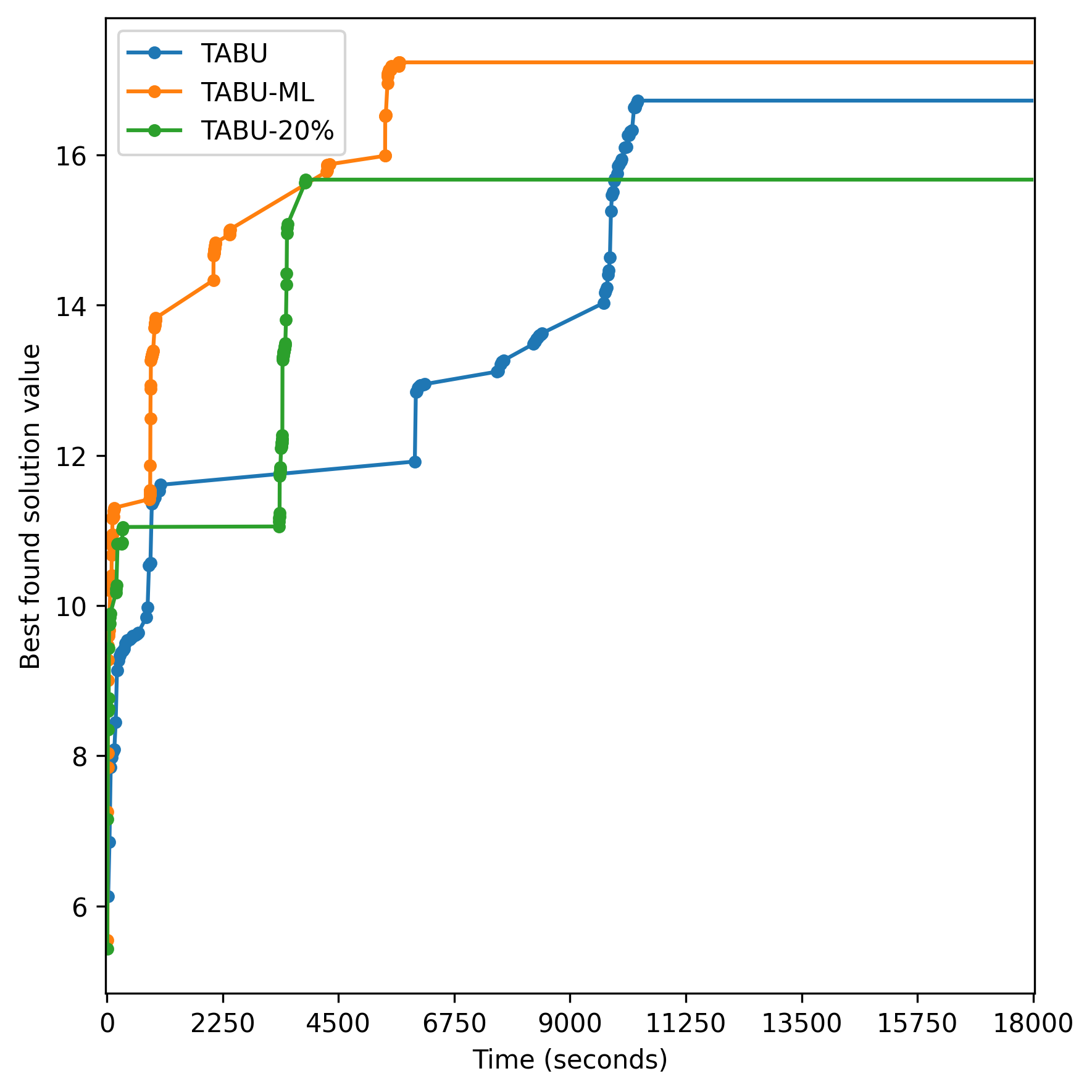}
    \caption*{50 nodes}
\end{subfigure}

\caption{Evolution of the best solution value over time for the three Tabu Search variants.}
\label{fig:tabu_evolution}
\end{figure}

To further analyze this behavior from a computational-time perspective, Figure~\ref{fig:tabu_evolution} presents the evolution of the best solution value over time for the three algorithms, considering the best execution of the first instance with 20, 30, and 50 nodes. These curves allow a direct comparison of the convergence behavior of the different approaches by showing how rapidly each algorithm improves the incumbent solution throughout the search process.

Overall, the results show that \texttt{TABU-ML} generally reaches high-quality solutions earlier than the two reference methods and achieves better final solution values in most cases. This behavior indicates that the learned classifier effectively guides the search toward more promising edge exchanges by prioritizing candidate moves that are more likely to lead to improvements. Consequently, \texttt{TABU-ML} reduces the computational effort spent evaluating less relevant neighborhood moves while accelerating the discovery of high-quality network topologies.


\section{Conclusion and future work}\label{sec:conclusion}

In this paper, we proposed an ML-guided Tabu Search approach for the tactical wireless network design problem. The motivation behind this work is the observation that the baseline \texttt{TABU} algorithm proposed in \cite{WH1} spends a significant amount of computational effort evaluating large neighborhoods. At each iteration, the search explores edge-exchange moves, in which one edge is removed from the current topology and another edge is added to reconnect the resulting components. Although this neighborhood structure is effective for improving the network topology, its exhaustive evaluation becomes increasingly expensive as the problem size grows. Indeed, a large number of candidate edge removals and reconnection edges must be considered, and each resulting neighboring topology requires a complete network configuration. This configuration includes selecting the master hub, determining the partition of its neighboring nodes, assigning communication channels and frequencies to the edges, and determining the active antenna beams and their orientations. Consequently, evaluating a single neighborhood requires not only examining many possible topology modifications but also solving a complex configuration problem for each candidate solution.

To alleviate this computational burden, we introduced a learning component into the neighborhood generation process. More specifically, a graph neural network classifier was trained to evaluate the candidate droppable edges of the current topology. The classifier receives as input a graph representation of the current solution, including node and edge features that describe the network structure and its associated configuration. For each candidate edge, the model predicts whether removing this edge is likely to lead to a promising neighboring solution.

The training data were generated from previous executions of the baseline Tabu Search algorithm. An edge removal was labeled as promising if removing this edge resulted in at least one feasible edge-exchange move leading to an improvement of the current solution value. Therefore, the classifier learns from the search decisions and outcomes produced by the original algorithm, allowing it to identify edge removals that are more likely to generate beneficial moves during future executions.

The trained classifier was subsequently integrated into the Tabu Search algorithm. Instead of evaluating the complete set of droppable edges at each iteration, \texttt{TABU-ML} first applies the classifier to identify a reduced subset of candidate edges that are considered more promising. The neighborhood is then generated and evaluated only from the moves associated with this selected subset.

It is important to note that the learning model does not replace the optimization procedure, but rather acts as a guidance mechanism within the existing Tabu Search framework. All major components of the original algorithm are preserved, including feasibility checks, tabu restrictions, the aspiration criterion, reconnection evaluation, and objective function computation. The role of the machine learning component is therefore limited to improving the efficiency of the neighborhood exploration by focusing the search on more promising candidate moves and reducing the computational effort spent on edge exchanges that are unlikely to improve the current solution.

The computational results demonstrate the effectiveness of the proposed learning-guided strategy. Under identical stopping criteria and time limits, \texttt{TABU-ML} generally achieves better average solution quality than the baseline \texttt{TABU} algorithm. This improvement is particularly significant because it is obtained without modifying the objective function, the feasibility conditions, or any of the fundamental components of the optimization procedure. The performance gain is instead a consequence of the way the neighborhood exploration is guided.

By reducing the number of candidate moves that need to be evaluated at each iteration, \texttt{TABU-ML} decreases the computational effort required for neighborhood exploration. Each iteration can therefore be completed more quickly, allowing the algorithm to perform a larger number of iterations and investigate more candidate topologies within the same computational budget. The classifier thus enables a more efficient allocation of the available computation time by focusing the search on the most promising regions of the solution space.

This reduction in computational effort is a key advantage of the proposed approach. In the baseline \texttt{TABU} algorithm, a significant portion of the available computation time can be spent evaluating candidate moves that ultimately do not contribute to improving the current solution. By using the classifier as a filtering mechanism, \texttt{TABU-ML} restricts the neighborhood evaluation to a smaller subset of edge modifications that are more likely to be beneficial.

As a result, the search process can advance more rapidly, improve the incumbent solution earlier during the execution, and increase its chances of discovering high-quality topologies before the time limit is reached. This behavior is confirmed by the solution evolution plots, which show that when the objective value is tracked as a function of computation time, \texttt{TABU-ML} generally reaches competitive solution values faster than the baseline \texttt{TABU} method.

Overall, this study demonstrates that machine learning can effectively serve as a complementary component within a metaheuristic algorithm. Rather than attempting to learn the complete optimization process or generate solutions from scratch, the objective is to assist the search procedure by improving the efficiency of a computationally expensive decision step. In the proposed approach, the classifier supports the Tabu Search algorithm by guiding the exploration of the neighborhood toward more promising candidate moves and reducing the number of edge exchanges that require a complete evaluation. This integration results in faster search iterations, enables the exploration of a larger number of candidate solutions within a fixed computational budget, and leads to improved solution quality on average.

Several directions can be considered for future work. First, the learning component could be extended to other stages of the edge-exchange move. In the current approach, the classifier is used to identify promising edges to remove from the current topology. A natural extension would be to also learn which reconnection edges are the most promising candidates to add after an edge removal. Such an approach would allow both components of the edge-exchange operation, namely edge removal and edge addition, to be guided by learning models.

However, preliminary experiments suggest that this extension is more challenging. The main difficulty comes from the much larger number of possible candidate edges to add compared with the number of edges that can be removed. Indeed, for a topology with $n$ nodes, only $n-1$ edges belong to the current tree and can therefore be considered for removal, whereas the number of possible reconnection edges is of order $O(n^2)$. Consequently, learning to rank or classify promising addition edges involves a significantly larger and more complex search space, which may require additional strategies to efficiently generate and evaluate candidate reconnection moves.

Another possible avenue for the use of machine learning is to accelerate other combinatorial decisions involved in the configuration of a network topology. For example, the selection of the master hub has a significant impact on the resulting network configuration and, consequently, on the objective value. Instead of relying solely on deterministic criteria or manually designed heuristics, a learning model could be trained to identify promising master hub candidates based on the structural characteristics of the topology and the specific features of the network instance. Such a model could help guide the configuration process toward more promising choices while reducing the computational effort required to evaluate alternative configurations.

A further opportunity for applying machine learning lies in the partitioning of the successors of the master hub into subsets. This step also constitutes a challenging combinatorial problem, as a large number of possible partitions may need to be considered and their quality can have a substantial impact on the final wireless configuration. Learning-based models could be developed to predict promising partitions or to prioritize the evaluation of the most relevant ones. Such an approach could further reduce the computational burden associated with the configuration phase while maintaining the ability to identify high-quality network designs.

More generally, future work could explore the use of machine learning to assist a wider range of topology configuration decisions, including master hub selection, subset partitioning, antenna configuration, channel assignment, and frequency selection. These decisions are currently handled through deterministic procedures or heuristic rules, but they involve complex interactions between the network topology, interference levels, communication constraints, and achievable throughput. By learning from previously evaluated configurations and their resulting performance, machine learning models could help identify promising configurations more efficiently and reduce the computational effort required during the optimization process.

Finally, future experiments should investigate the generalization capabilities of the trained models across different instance sizes and network characteristics. In particular, it would be valuable to assess whether a model trained on smaller network instances can effectively guide the search process on larger instances, or whether dedicated approaches, such as transfer learning or model adaptation, are required to maintain performance when the characteristics of the problem instances change.\\

\noindent\textbf{Declaration of competing interest}\\
The authors declare that they have no competing interests. No funding was received from any organization that could be perceived as influencing the research presented in this paper.\\

\noindent\textbf{Data availability}\\
The data used for testing in this study are the same as those generated by \cite{WH1}.\\

\end{document}